\documentclass{article}

\usepackage{PRIMEarxiv}

\usepackage[utf8]{inputenc} 
\usepackage[T1]{fontenc}    
\usepackage{hyperref}       
\hypersetup{
    colorlinks=false,
    linkcolor=black,
    filecolor=black,      
    urlcolor=black,
    pdfpagemode=FullScreen,
}

\usepackage{url}            
\usepackage{booktabs}       
\usepackage{amsfonts}       
\usepackage{nicefrac}       
\usepackage{microtype}      
\usepackage{lipsum}
\usepackage{fancyhdr}       
\usepackage{graphicx}       
\graphicspath{{media/}}     
\usepackage{amsthm}
\newtheorem{definition}{Definition}
\usepackage{mathtools, nccmath}
\usepackage{amssymb, amsthm, mathrsfs}
\usepackage[table,xcdraw]{xcolor}

\newtheorem{prop}{Proposition}
\usepackage{float}
\restylefloat{table}
\title{Image Classification using Combination of Topological Features and Neural Networks
}

\author{
  Mariana Dória Prata Lima,  Gilson Antônio Giraldi \\
  Laboratório Nacional de Computação Científica \\
  Petrópolis, RJ, Brasil\\
   \And
  Gastão Florêncio Miranda Junior \\
  Universidade Federal de Sergipe \\
  São Cristóvão, SE, Brasil\\
}

\begin{document}
\maketitle

\begin{abstract}
In this work we use the persistent homology method, a technique in topological data analysis (TDA), to extract essential topological features from the data space and combine them with deep learning features for classification tasks. In TDA, the concepts of complexes and filtration are building blocks. Firstly, a filtration is constructed from some complex. Then, persistent homology classes are computed, and their evolution along the filtration is visualized through the persistence diagram. 
Additionally, we applied vectorization techniques to the persistence diagram to make this topological information compatible with machine learning algorithms. This was carried out with the aim of classifying images from multiple classes in the MNIST dataset. Our approach inserts topological features into deep learning approaches composed by single and two-streams neural networks architectures based on a multi-layer perceptron (MLP) and a convolutional neral network (CNN) taylored for multi-class classification in the MNIST dataset.
In our analysis, we evaluated the obtained results and compared them with the outcomes achieved through the baselines that are available in the TensorFlow library.  The main conclusion is that topological information may increase neural network accuracy in multi-class classification tasks with the price of computational complexity of persistent homology calculation. Up to the best of our knowledge, it is the first work that combines deep learning features and the combination of topological features for multi-class classification tasks.
\end{abstract}

\keywords{topological data analysis \and neural network \and images \and persistence homology \and classification}

\section{Introduction}

Topological Data Analysis (TDA) is a multidisciplinary field dedicated to the study of qualitative and quantitative features of complex datasets \cite{Munch_2017}. A prominent approach in this field is persistent homology (PH), which combines principles from algebraic topology and computational geometry to identify and quantify topological features such as connected components and holes of different dimensions \cite{Bubenik2012StatisticalTD}. This allows for the detection of transient and persistent structural patterns in data, providing essential insights into its properties.

Currently, studies utilizing this approach are at the forefront of research in an increasingly diverse set of applications, including neuroscience \cite{3, 4}, disease detection \cite{2, 10}, analysis of electric motor data \cite{11}, fluorescence microscopy images \cite{12}, satellite images \cite{13, 16} proteins \cite{17}, natural language processing \cite{18, 19}, among others \cite{31, 32}.

The usual pipeline in TDA involves the following steps: (a) Computation of a simplicial or cubical complex $K$; (b) Filtration of  $K$; (b) Calculation of PH groups over the filtration; (c) Tracking of the evolution of PH classes in the filtration; (d) Persistence diagram (PD) construction. 
In this pipeline, one of the cornerstones is the representation of the topological features present in the data through a PD. This diagram not only identifies these features but also evaluates their relevance and persistence throughout the data space. However, it is crucial to vectorize the PD to translate the topological information into formats compatible with machine learning algorithms \cite{Leykam2022TopologicalDA}, statistical analysis \cite{Bubenik2012StatisticalTD}, and other data processing techniques. This task has been tackled through distance measures, like Wasserstein and bottleneck distances \cite{Park2021UnsupervisedLO,Tirelli2021LearningQP}, in persistence diagrams spaces; and vectorization approaches that project PDs into real vector spaces \cite{Ali2022ASO}.

There are several approaches to vectorizing the PD, namely Betti curve \cite{33, 11, 38}, persistence landscape \cite{26}, persistence silhouette (PS) \cite{30}, persistence image (PI) \cite{34, 36, 37} and heat kernel (HK) \cite{35},  each with its advantages and limitations as highlighted in related works.
For example, reference \cite{38} explores methods involving Betti curve, persistence landscape, and persistence silhouette vectorization to classify rough surfaces using traditional classification methods such as polynomial radial base Support Vector Machine (SVM) and sigmoid kernel SVM, decision trees, random forests, multi-layer perceptrons, as well as k-nearest neighbors voting.

The performance obtained is compared with conventional surface roughness parameters based on areas. When using the Betti curve as input, the accuracy exceeds that achieved by classifiers using conventional parameters. However, in the case of the persistence landscape and persistence silhouette, while the accuracy also surpasses that of classifiers using conventional parameters, the results are slightly lower compared to those obtained using the Betti curve.
In \cite{1}, the pipeline (a)-(d) is applied and several PD vectorization methods are applied. Then, the approach employs combinations of topological features that are used as inputs for SVM, RF and Lasso machine learning methods serving as classifiers for mult-class tasks in MNIST images.

On the other hand, the reference \cite{2} proposes the TDA-Net, composed by single and two-stream deep learning approaches formed by convolutional neural network (CNN) and multi-layer perceptron (MLP) architectures that fuses vectorized PDs representations and deep features for distinguishing between negative and COVID positive samples in Chest X-ray datasets. The main goal of those works is to obtain improved accuracies and perform result analysis.

We follow \cite{1,2} and consider the mentioned  persistence diagram vectorization methods for image classification focusing on integrating combinations of topological features into neural networks for image classification. Likewise in \cite{1}, the pipeline for feature extraction has eight steps: (1) Binarization of images; (2) Binary image processing; (3) Computation of a cubical complex $\mathcal{X}$; (4) Filtration of  $\mathcal{X}$; (5) Calculation of cubical PH over the filtration; (6) PD construction; (7) PD vectorization. However, differently from \cite{1} that uses traditional classifiers over MNIST data, we have applied deep neural networks architectures constructed using two neural networks tailored for MNIST data classification named MNIST-CNN and MNIST-MLP. Our deep leaning approach is inspired in the one proposed in \cite{2} but, differently from that reference, we tackle a multi-class problem and apply the MNIST-CNN and MNIST-MLP from Tensorflow \cite{21}, which is a library widely used for machine learning and deep learning model development.   

The considered feature spaces are formed by topological features of dimensions zero and one, their concatenation and fusion (see section \ref{sec:proposedmethod}). In this context, the features spaces generated through the topological features are used for image classification with emphasis on images from the MNIST, fully recognized in the machine learning community. We apply the considered neural networks as follows: (1) MNIST-CNN and MNIST-MLP alone, applied as as baselines, where the former receives the original image as imput while the latter processes the original data in vectorized form; (2) MNIST-MLP alone receiving the topological features; (3) Two-stream composed by two MNIST-MLPs with the first one processing the topological features and the other one receiveing the original vectorized image as input; (4) Two-stream composed by MNIST-MLP and MNIST-CNN with the former processing the topological features and the latter processing the original image.

We evaluate the obtained results by recording the accuracy of the  different feature spaces and architectures (1)-(4) on the MNIST dataset. We conclude that topological information increases the accuracy of the MLP neural network in multi-class classification tasks. However, in the case of CNN, we did not observe performance improvement compared with the baseline receiving the original images as input.
In our implementation, we make use of the \textit{giotto-tda} library \cite{20} to compute topological features. The neural networks are implemented through Tensorflow library \cite{21}.

The contributions of this work are: 
\begin{itemize}
    \item The application of topological features coding in different architectures of neural networks for multi-class image classification from the MNIST dataset.
    \item Computational experiments using a variety of PD vectorization methods and the considered deep learning architectures and their combination on the MNIST dataset.
    \item Improviment in the MNIST MLP accuracy compared with the baseline receiving only the original images as input.
\end{itemize}

Up to the best of our knowledge, it is the first work that combines deep learning features and topological features for a multi-class classification tasks.

The structure of this paper is as follows. We begin, in Section \ref{sec:mathematicalbackground}, introducing the reader to some concepts of algebraic topology that will be used throughout the work. Furthermore, we present several three functions that recalculate the intensity of pixels in images from binary images. In Section \ref{sec:proposedmethod}, we explain how to extract machine learning-ready features from the obtained persistence diagrams and describe a generic TDA machine learning pipeline. In Section \ref{sec:results}, we present the results and interpret the features that were important for classifying MNIST images. Finally, we discuss the conclusions in Section \ref{sec:conclusions}.

\section{Mathematical Background}
\label{sec:mathematicalbackground}
In this section, we will introduce the fundamental mathematical principles necessary to understand the proposed methodology. We will discuss key concepts, including cubical complexes \cite{23}, homology groups, complex filtration for the extraction of topological features \cite{22, 23}, graph representations \cite{22, 24}, and the image processing steps performed prior to the extraction of topological features \cite{1, 30}.

\subsection{Cubical Complex}
\label{subsec:cubical-complex}

In the context of persistent homology, cubical complexes are a common choice for representing spatial data, such as images. Before defining what a cubical complex is, it's important to establish some terminology.
\\
\begin{definition}[Elementary Interval] An \textit{elementary interval} is a closed interval $ I \subset \mathbb{R}$ of the form
\begin{equation}
    I = [l, l+1] \; or \; I = [l, l] = \{l\}, \label{eq:element-interval}  
\end{equation}
for some $l \in \mathbb{Z}$. These two forms are called respectively non-degenerate and degenerate. 

\end{definition}

\begin{definition}[Elementary Cube] An elementary cube $\mathcal{Q}$ is a finite product of elementary intervals of the form

    \begin{equation}
        \mathcal{Q} = I_1 \times I_2 \times \dots \times I_d \subset \mathbb{R}^d, \label{element-cube}
    \end{equation}
where each $I_i, i=1, 2, \dots, d$ is an elementary interval.  The set of all elementary cubes in $\mathbb{R}^d$ is denoted by $\mathcal{K}^d$. The set of all elementary cubes is denoted by $\mathcal{K}$, namely

\begin{equation}
    \mathcal{K} =  \bigcup_{d=1}^{\infty}{\mathcal{K}^d}. \label{set-elementary-cubes}
\end{equation}
    
\end{definition}

\begin{definition}[Component of elementary cube] Let $\mathcal{Q} = I_1 \times I_2 \times \dots \times I_d \subset \mathbb{R}^d$ be an elementary cube. The interval $I_i$ is referred to as $i$th component of $\mathcal{Q}$ and is written as $I_i(\mathcal{Q}).$

\end{definition}

\begin{definition}[Dimension of elementary cube] Consider $\mathcal{Q} = I_1 \times I_2 \times \dots \times I_d \subset \mathbb{R}^d$ be an cube. The dimension of $\mathcal{Q}$ is defined to be the number of nondegenerate components in $\mathcal{Q}$ and is denoted by $dim \; \mathcal{Q}$

\end{definition}

\begin{definition}[Embedding number of elementary cube]
    Let $Q = I_1 \times I_2 \times \dots \times I_d \subset \mathbb{R}^d$ be an elementary cube. The embedding number of $Q$ is denoted by $emb \; Q$ and is defined to be d since $Q \subset \mathbb{R}^d$.
\end{definition}

\begin{definition}[Face] Let $Q, P \in \mathcal{K}$. If $Q \subseteq P$, then $Q$ is a face of $P$. If $Q \subset P$, then $Q$ is proper face of $P$. $Q$ is primary face of $P$, if $Q$ is a face of $P$ and $dim \;Q = dim \; P - 1$.
\end{definition}

\begin{definition}[Cubical Complex] A cubical complex $\mathcal{X} \subset \mathbb{R}^d$ is defined by a finite union of elementary cubes.
If $\mathcal{X}$ is cubical complex then we denote the set of all elementary cubes in $\mathcal{X}$ by
\begin{center}
    $\mathcal{K}(\mathcal{X}) = \{ Q \in \mathcal{K} | Q \subset \mathcal{X}\}$
    
\end{center}
 and all elementary cubes of $\mathcal{X}$ of dimension $k$ is denoted by $\mathcal{K}_k(\mathcal{X}) = \{Q \in \mathcal{K}(\mathcal{X}) | dim \; Q = k\}$. The elements of the latter are called by $k$-cubes of $X$.   
\end{definition}

\begin{figure}
  \centering
  \includegraphics{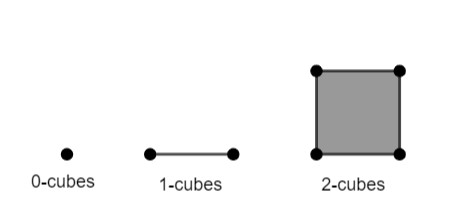}
  \caption{Cubes of dimension 0, 1 and 2.}
  \label{fig:fig1}
\end{figure}

Thus the cubical complexes are constructed from cubes of various dimensions. We begin with dimension 0-cubes, which correspond to the vertices of the structure, typically associated with image pixels. Next, we have dimension 1-cubes representing the edges, and dimension 2-cubes representing the faces, as shown in Figure \ref{fig:fig1}. Although this structure can be extended to cubes of higher dimensions, in this work, we will restrict ourselves to cubes up to dimension 2. For more details on cubical complexes, please refer to \cite{23}%

\subsection{Cubical Homology}
\label{sec:cubical-homology}
Homology is a tool that combines topological, geometric, and algebraic principles to comprehend and quantify the topological and structural properties of geometric objects in multidimensional spaces \cite{22}. At the core of homology are homology groups, algebraic structures that represent the topological characteristics of an object or space. The construction of these groups is based on fundamental concepts, such as cycles and boundaries. However, to grasp these concepts, it is necessary to establish basic definitions, such as $d$-chain and chain complex.

\begin{definition}
    The set of elementary cubes in $\mathbb{R}^d$ of dimension $k$ is denoted by 
    \begin{equation}
        \mathcal{K}_k^d = \mathcal{K}^d \cap \mathcal{K}_k,
     \end{equation}
where $\mathcal{K}_k := \{Q \in \mathcal{K}| dim \; Q = k\}$.

\end{definition} 

\begin{definition}[$k$-chain]
    With each $k$-cube $Q \in \mathcal{K}_k^d$ , we associate an algebraic object $\widehat{Q}$ called an elementary $k$-chain of $\mathbb{R}^d$. The set of all elementary $k$-chains of $\mathbb{R}^d$ is given by 
    \begin{equation}
        \widehat{\mathcal{K}_k^d} = \{\widehat{Q} | Q \in \mathcal{K}_k^d \}.
    \end{equation}
     Given any finite collection $\{\widehat{Q}_1, \dots, \widehat{Q}_m\} \subset \mathcal{K}_k^d$ of $k$-dimensional elementary chains, we are allowed to consider sums of the form
    \begin{equation}
    c = \alpha_1 \widehat{Q}_1 + \alpha_2 \widehat{Q}_2 + \dots + \alpha_m \widehat{Q}_m,
    \end{equation}
    called $k$-chains, the set of which is denoted by $C_k^d$.
\end{definition}

\begin{definition}[Set of elementary $k$-chains] The set of elementary $k$-chains of a cubical complex $\mathcal{X}$ is denoted by 

\begin{equation}
    \widehat{\mathcal{K}}_k(\mathcal{X}) = \{ \widehat{Q}| Q \in \mathcal{K}_k(\mathcal{X}) \}. 
\end{equation}    
\end{definition}

\begin{definition}[Set of $k$-chains of a cubical complex] Let $\mathcal{X} \subset \mathbb{R}^d $ a cubical complex. The set of $k$-chains of $\mathcal{X}$ constructed by finite number of elements of $\widehat{\mathcal{K}}_k(\mathcal{X})$ is denoted by $C_k(\mathcal{X})$.
    
\end{definition}

\begin{definition}[Addition operation of $k$-chains]  Consider $c_1, c_2 \in C_k(\mathcal{X})$, where $c_1 = \sum_{i=1}^{m} {\alpha_i \widehat{Q}_i}$ and $c_2 = \sum_{i=1}^{m} {\beta_i \widehat{Q}_i}$, where $\alpha_i, \beta_i \in \{0, 1\}$ with $i=1, 2, \dots, m$. The addition of $c_1$ and $c_2$ is defined by

\begin{equation}
    c = c_1 + c_2 =  \sum_{i=1}^{m} {\alpha_i \widehat{Q}_i} +  \sum_{i=1}^{m} {\beta_i \widehat{Q}_i} = \sum_{i=1}^{m} (\alpha_i + \beta_i) \widehat{Q}_i,
\label{eq:sum-of-chains}
\end{equation}
where all additions ($ \alpha_i + \beta_i $) are module 2.
\end{definition}

\begin{prop}[Group of $k$-chains] The set $C_k(\mathcal{X})$  with addition operation is an abelian group.
    
\end{prop}
Proof: See \cite{23}.

If $Q \subset \mathcal{X} \subset \mathbb{R}^d$ then $emb \; Q = d$. This in turn implies that $Q \in \mathcal{K}^d$, so to use the notation $\mathcal{K}^d(\mathcal{X})$ is somewhat redundant, but it serves to remind us that $\mathcal{X}\subset \mathbb{R}^d$. Therefore, when it is convenient we will write $\mathcal{K}^d(\mathcal{X})$ and also $\mathcal{K}_k^d(\mathcal{X}) := \mathcal{K}^d(\mathcal{X})\cap \mathcal{K}_k(\mathcal{X})$.The elements of $\mathcal{K}_0(\mathcal{X})$ are the vertices of $\mathcal{X}$ and the elements of $\mathcal{K}_1(\mathcal{X})$ are the edges of $\mathcal{X}$. More generally, the elements of $\mathcal{K}_k(\mathcal{X})$ are the $k$-cubes of $\mathcal{X}$.
Now we will define the boundary operator, but to do so we need to define some propositions that will not be proven in this article. The reader can find them at \cite{23} also.

\begin{definition}
    Consider $c_1, c_2 \in C_k^d$, where $c_1 = \sum_{i=1}^{m} {\alpha_i \widehat{Q}_i}$ and $c_2 = \sum_{i=1}^{m} {\beta_i \widehat{Q}_i}$. The scalar product of the chains $c_1$ and $c_2$ is defined as

    \begin{equation}
        \langle c_1, c_2 \rangle := \sum_{i=1}^{m}{\alpha_i}{\beta_i}.
    \end{equation}
\end{definition}

\begin{prop} \label{prop1} Given two elementary cubes $P \in K^d_k$ and $Q \in K^{d'}_{k'}$ set

\begin{equation}
    \widehat{P} \; \star\; \widehat{Q} := \widehat{P \times Q},
\end{equation}
where $\widehat{P \times Q}$ is algebraic representation of $P \times Q$.
This definition can be extended to arbitrary chains $c_1 \in C^d_k$ and $c_2 \in C^{d'}_{k'}$ by

\begin{equation}
   c_1 \; \star \;  c_2 := \sum_{P\in \mathcal{K}_k, Q \in \mathcal{K}_{k'}} \langle c_1, \widehat{P} \rangle \langle c_2, \widehat{Q} \rangle \widehat{P \times Q},
\end{equation}

The chain $c_1 \; \star \;  c_2 \in C^{d+d'}_{k+k'}$ is called the cubical product of $c_1$ and $c_2$. 
\end{prop}

\begin{prop} \label{prop3} Let $\widehat{Q}$ be an elementary cubical chain of $\mathbb{R}^d$ with $d > 1$. Then there exist unique elementary cubical chains $\widehat{I}$ and $\widehat{P}$ with $emb(I) = 1$ and $emb(P) = d - 1$ such that
\begin{equation}
    \widehat{Q} = \widehat{I} \; \star \; \widehat{P}.
\end{equation}
\end{prop}

\begin{definition}[Boundary Operator] 

For each $k \in \mathbb{Z}$ the cubical boundary operator $C_{k}^d$.
\begin{equation}
    \partial_k \colon C_{k}^d \rightarrow C_{k-1}^d,
\end{equation} defined for an elementary chain $\widehat{Q} \in \widehat{\mathcal{K}}_k^d$ by induction on the embedding number $d$ as follows.

For the case $d = 1$. Then $Q$ is an elementary interval and hence $Q = [l] \in \mathcal{K}_0^1 $ or $Q = [l,l + 1] \in \mathcal{K}_1^1$ for some $l \in \mathbb{Z}$. Define

\begin{equation}
     \partial_k \widehat{Q} :=
    \begin{cases}
          0, & \textit{if } Q = [l]\\
          \widehat{[l + 1]} - \widehat{[l]}, &\textit{if } Q = [l, l + 1].
\end{cases}
\end{equation}

Now assume that $d > 1$. Let $I = I_1(Q)$ and $P = I_2(Q) \times \dots \times I_d(Q)$, where $I_i(Q)$ is the $i$th component of $Q$. Then
by Proposition \ref{prop3},
\begin{center}
    $\widehat{Q} = \widehat{I} \; \star \; \widehat{P}$.
\end{center}

Define:

\begin{equation}
    \partial_k \widehat{Q} := \partial_{k_1} \widehat{I} \; \star \; \widehat{P} + (-1)^{\text{dim I}} \widehat{I} \; \star \; \partial_{k_2} \widehat{P},
\end{equation}
where $k_1 = dim \; I$, $k_2 = dim \; P$. Finally, we extend the definition to all chains by linearity; that is, if $c = \alpha_1\widehat{Q}_1 + \alpha_2 \widehat{Q}_2 + \dots + \alpha_m \widehat{Q}_m$, then

\begin{center}
    $\partial_k c := \alpha_1 \partial_k \widehat{Q}_1 + \alpha_2 \partial_k \widehat{Q}_2 + \dots + \alpha_m \partial_k \widehat{Q}_m$.
\end{center}
\end{definition}

\begin{prop} \label{prop4} $\partial \circ \partial = 0$.
    
\end{prop}
Proof: See \cite{23}

\begin{definition}[$k$-cycle]
Let $\mathcal{X}$ a cubical complex. A $k$-chain, denoted as $z \in C_k(\mathcal{X})$, is a $k$-cycle within $\mathcal{X}$ if it satisfies the condition $\partial z = 0$
    
\end{definition}

\begin{definition} [Set of all $k$-cycle]
The kernel of a linear map refers to the set of elements that are mapped to zero, and this forms a subgroup of the domain.  Thus the set of all $k$-cycles
in $\mathcal{X}$, which is denoted by $Z_k(\mathcal{X})$, is $ker\left( \partial_k \right)$ and forms a subgroup of $C_k(\mathcal{X})$. To formally express this relationship:
\begin{equation}
Z_k(\mathcal{X}) := ker\left( \partial_k \right) = C_k(\mathcal{X}) \cap ker\left( \partial_k \right) \subset C_k(\mathcal{X}) = \{c \in C_k(\mathcal{X}) | \partial_k c = 0\}
\end{equation}
    
\end{definition}

\begin{definition}[$k$-boundary]
A $k$-chain $z \in C_k(\mathcal{X})$ is labeled as a $k$-boundary within $\mathcal{X}$ if there exists another $(k+1)$-chain, denoted as $c \in C_{k+1}(\mathcal{X})$, such that $\partial c = z$.

\end{definition}

\begin{definition}[Set of all $k$-boundary]
The set of boundary elements in $C_k(\mathcal{X})$, denoted as $B_k(\mathcal{X})$, consists of the image of the boundary operator $\partial_{k+1}$. Since $\partial_{k+1}$ is a homomorphism, $B_k(\mathcal{X})$ forms a subgroup of $C_k(\mathcal{X})$.

This relationship can be summarized as follows:

\begin{equation}
B_k(\mathcal{X}) := \text{im} \partial_{k+1} = \partial_{k+1}(C_{k+1}(\mathcal{X})) \subset C_k(\mathcal{X}) = \{c \in C_{k}(\mathcal{X}) | \partial_{k+1}(d) = c \; \text{with} \; d \in C_{k+1}(\mathcal{X})\}.
\end{equation}

\end{definition}

It's important to note that, as demonstrated by Proposition \ref{prop4}, if $\partial c = z$, then $\partial z = \partial^2 c = 0$. Consequently, every boundary is also a cycle, and as a result, $B_k(\mathcal{X})$ is a subgroup of $Z_k(\mathcal{X})$.

\begin{definition}
    The $k$th cubical homology group, or briefly the $k$th homology group of $\mathcal{X}$, is the quotient group
    \begin{equation}
        H_k(\mathcal{X}) := Z_k(\mathcal{X})/B_k(\mathcal{X}) = \{ c + B_k(\mathcal{X})| c \in Z_k(\mathcal{X})\}
        \label{eq:homology-group}
    \end{equation}
    The dimension of homology group is called Betti Number: $dim(H_k(\mathcal{X})) = \beta_k$.
\end{definition}

Homology groups play a fundamental role in identifying holes, voids, and other topological features in complex objects. An extension of homology is persistent homology defined in section \ref{sec:persistence-homology}. Specifically, the Betti number $\beta_0$ is the number of connected component, the Betti number $\beta_1$ is the number of holes and $\beta_k$ is the number of voids of dimension $k$ with $k > 1$.

\subsection{Persistence Homology}
\label{sec:persistence-homology}
Persistent homology is a field that investigates topological invariants and enables the extraction of valuable topological information, including the counting of connected components and the enumeration of voids in the data space \cite{22}.

While this tool is commonly applied to point clouds, it can also be extended to images. In an image, data is represented by pixels or voxels. While it is conceivable to consider the pixels of an image as vertices in certain structures, such as the Vietoris-Rips simplicial complex,  not all pixels need to be directly connected, implying the utilization of vertices based on specific intensities.

To analyze images effectively, a common approach is to represent the pixels as $0$-cubes on a lattice. By doing so, we can apply cubical homology (section \ref{sec:cubical-homology}) to uncover underlying patterns in the images. This process involves cubical complexes generation, as defined in section \ref{subsec:cubical-complex}, and filtration defined next.


\subsubsection{Filtration}
\label{subsec:filtration}
Filtration is essentially an ordered sequence of topological spaces, where, as you progress through this sequence, more details or information are added. With this progressive process, it becomes possible to track the evolution of the topological characteristics of the data. Formally, we can define it as follows:

\begin{definition}[Filtration]
    
Let $\mathbf{x}\in\mathbb{R}^{n\times m}$ a grayscale image and
$G = \left\{ \begin{array}{cccc}
0, & 1, & \ldots & ,n-1\end{array}\right\} \times\left\{ \begin{array}{cccc}
0, & 1, & \ldots & ,m-1\end{array}\right\} \in\mathbb{R}^{2}$ the corresponding grid. In this case, each point (pixel) $\mathbf{p}\in G$
is 0-cube in $\mathbb{R}^{2}$, each edge $\left(\mathbf{p},\mathbf{q}\right)\in G \times G$
is a 1-cube, etc. Also, we consider a cubic complex $\mathcal{X}$
built over subset of $G$ and a function $f:\mathcal{K}_{0}(\mathcal{X})\longrightarrow\mathbb{R_{+}}$.
So, we can assembly a set $S=\left\{ \mathbf{p}_{1},\mathbf{p}_{2},\ldots,\mathbf{p}_{N}\right\} \subset\mathcal{K}_{0}(\mathcal{X})$,
such that:

\begin{equation}
f\left(\mathbf{p}_{1}\right) \leq f\left(\mathbf{p}_{2}\right) \leq \ldots \leq f\left(\mathbf{p}_{N}\right),\label{eq:ordered-points}
\end{equation}
If $a_{i}=f\left(\mathbf{p}_{i}\right)$, $\mathcal{X}_{0}=\emptyset$, and
$\mathcal{X}_{i}$ is the subcomplex defined by the subset $\left\{ \mathbf{q}\in\mathcal{K}_{0}(\mathcal{X});f\left(\mathbf{q}\right)\leq a_{i}\right\} $
then we call the sequence $\mathcal{X}_{0},\mathcal{X}_{1},\mathcal{X}_{2},\ldots,\mathcal{X}_{N}$
a filtration of $\mathcal{X}$ if it satisfies :

\begin{equation}
\emptyset=\mathcal{X}_{0}\subset\mathcal{X}_{1}\subset\dots\subset\mathcal{X}_{N}=\mathcal{X}.\label{eq:nested-sequence-cubic}
\end{equation}

The relation $\mathcal{X}_{i-1} \subset \mathcal{X}_i$ induces a homomorphism between  homology groups: 
\begin{equation}
f_d^{i-1,i} \colon H_d(\mathcal{X}_{i-1}) \rightarrow H_d(\mathcal{X}_i). \label{eq:indeced-homomorph}    
\end{equation}

The nested sequence of complexes corresponds, therefore, to sequences of homology groups connected by homomorphisms: $0 = H_d(\mathcal{X}_0) \rightarrow H_d(\mathcal{X}_1) \rightarrow \dots \rightarrow H_d(\mathcal{X}_n) = H_d(\mathcal{X})$, for each dimension $d$. As $\mathcal{X}_{i-1}$ progresses to $\mathcal{X}_i$, new homology classes can be formed or lost or merge with others \cite{22}. 
\end{definition}



As we proceed forward in the sequence (\ref{eq:nested-sequence-cubic}), new topological elements begin to form (or "birth"), while some of these elements merge (or "die") as the filtration continues. The instances of birth and death can be graphically depicted using \textit{diagrams}, which will be formally defined later in this subsection. The straightforward choice for the function $f:\mathcal{K}_{0}(\mathcal{X})\longrightarrow\mathbb{R_{+}}$ is the gray-level image intensity. Other possibilities will be seen following.

\subsubsection{Persistence Homology Group}
\label{subsubsec:persistence-homology-group}
Consider $\mathcal{X}$ a cubical complex, the sequence (\ref{eq:nested-sequence-cubic}) and the subcomplexes $\mathcal{X}_i, \mathcal{X}_j \subseteq \mathcal{X}$. For each $0 \leq i \leq j \leq n$, we have an inclusion map from the underlying space of $\mathcal{X}_i$ to that of $\mathcal{X}_j$, and thus an induced homomorphism $f_d^{i,j} \colon H_d(\mathcal{X}_i) \rightarrow H_p(\mathcal{X}_j)$ for each dimension $d$. The \textit{$
d$-dimensional persistent homology group}, denoted as $H_d^{i,j}$, is the image of the induced homomorphism, or $\text{im} f_d^{i,j}$ \cite{22}.

\subsubsection{Persistence and Barcode Diagrams}
The homology class $u \in H_d(\mathcal{X}_i)$ born in $\mathcal{X}_i$ if $u \notin H_{d}^{i-1, i}$ and dies in $\mathcal{X}_j$ if $f_d^{i,j-1}(u) \notin H_{d}^{i-1, j-1}$ but $f_d^{i,j}(u) \in H_d^{i-1, j}$. The persistence diagram and the barcode diagram are representations of the interval $[i,j]$, where $i$ is the birth time of an homology class $u$, and $j$ is its death time, with $j > i$.  The persistence diagram represents each interval $[i,j]$  as a point $(i, j)$ in the plane $xy$, while the barcode diagram represents the interval through a bar.

\begin{figure}
  \centering
  \includegraphics[scale=0.6]{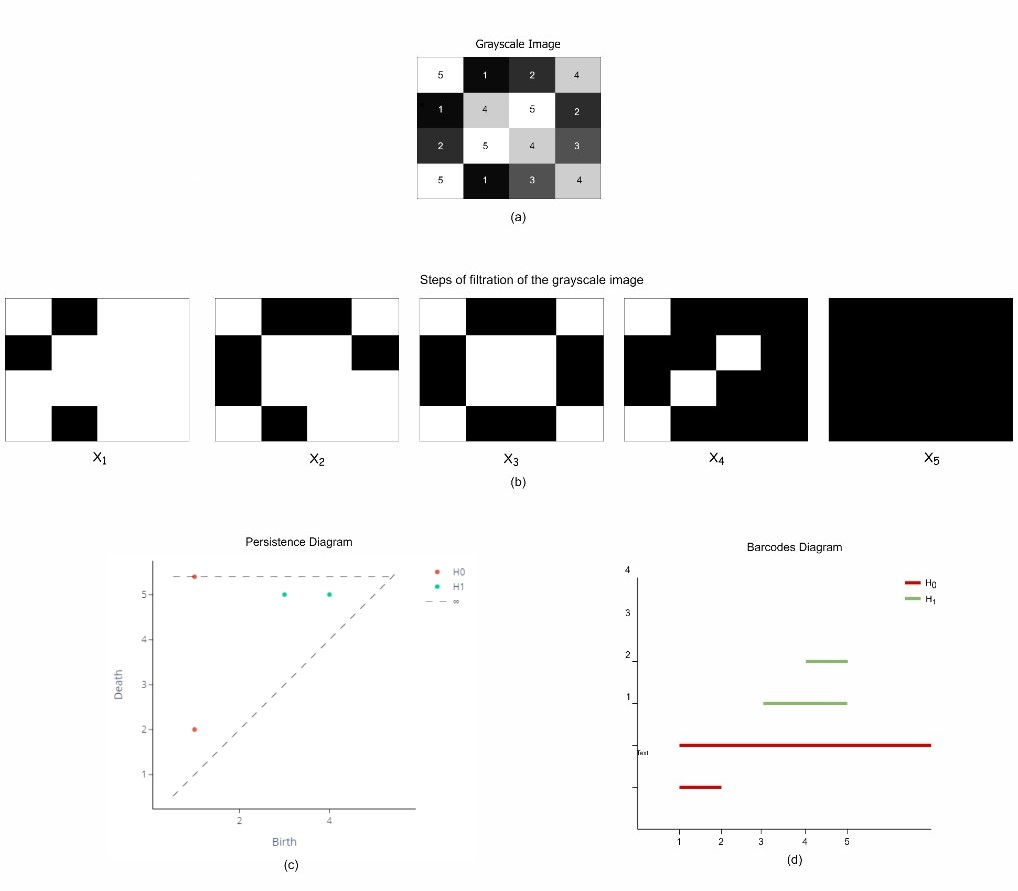}
  \caption{(a) Example of grayscale image; (b) Filtration of a grayscale image and (c) its persistence diagram and (d) its barcodes diagram.}
  \label{fig:fig2}
\end{figure}

As an example, let us consider Figure \ref{fig:fig2} that shows an grayscale image (a). The corresponding filtration with respect to the grayscale intensities is picture on Figure  \ref{fig:fig2}.(b). Then we build persistence homology groups of dimensions $d=0$ and $d=1$, whose births and deaths of persistence homology classes are represented in the persistence diagram of Figure \ref{fig:fig2}.(c). The barcode diagram of these persistence homology classes are represented in the Figure \ref{fig:fig2}.(d). Each coordinate of a point $(i,j)$, in a persistence diagram is associated with a multiplicity, represented as $\mu(d_i)$, which indicates how many connected components (if $d = 0$) or d-dimensional holes (if $d \geq 0$) share the same moment of birth and death \cite{22, 24}.

\subsubsection{Representation of Persistence Diagram}

The representation of persistence diagrams plays a fundamental role in translating the information therein into formats compatible with machine learning algorithms. In this section, we will explore the main methods employed to achieve this representation of persistence diagrams,  including Betti curve \cite{29}, persistence landscape \cite{26, 27}, power-weighted silhouette \cite{27}, persistence image \cite{25} and heat kernel \cite{28}.\\
\\
Let a $d$ the dimension of persistence homology group and $\mathcal{D} = \{(b_i, d_i)\}_{i=1}^n$ its persistence diagram.

\begin{enumerate}
    \item  Betti curve:  Consider the function $\mathcal{B}_d : \mathbb{R} \longrightarrow \mathbb{N}$ that, for $\varepsilon \in \mathbb{R}$, returns the number of intervals that have $b_i$ as lower and $d_i$ as upper bounds and that contains $\varepsilon$, that is:
    \begin{align*}
        \mathcal{B}_d \colon \mathbb{R} & \longrightarrow \mathbb{N}\\
        \varepsilon&\longmapsto |\{(b_i, d_i) ; \quad \varepsilon \in [b_i, d_i) \}|.
    \end{align*}
   The curve that represents the function $\mathcal{B}_n$ is named Betti curve.
   
    \item Persistence Landscape:
      The persistence landscape is the family $\{\lambda_k\}_{k \in \mathbb{N}}$ of  functions $\lambda_k \colon \mathbb{R} \longrightarrow [0, \infty)$, where $\lambda_k(t)$ is the $k$-th largest value of the set $\{\Lambda_i(t)\}_{i=1}^n$ and
    
    \begin{center}
    \[ \Lambda_i(t) = 
    \begin{cases}
    t - b_i,& \text{if }\; t \in [b_i, \frac{b_i + d_i}{2}],\\
    t + b_i, &\text{if}\; t \in (\frac{b_i + d_i}{2}, d_i],\\
    0,              & \text{otherwise},
\end{cases}
\]
    \end{center}
 where $i = 0, 1, \dots, n$ and the parameter $k$ is called the layer. The function is then vectorized on a uniform grid with resolution $r \times r$.

    \item Power-Weighted Silhouette:
    Consider $\mathcal{D} = \{(b_i, d_i)\}_{i=1}^n$ a persistence diagram with $m$ off diagonal points.  The power-weighted silhouette is the weighted average of the functions $\Lambda_i(t)$ defined by
    \begin{center}
        $\phi_p(t) = \frac{\sum_{j = 1}^{m}{|d_j - b_j|^p \Lambda_j(t)}}{\sum_{j = 1}^{m}{|d_j - b_j|^p}}$
    \end{center}
    The parameter $p$ can be seen as a trade-off value between uniformly treating all pairs in the persistence diagram or considering only the most persistent pairs. Therefore, when $p$ is small, the function $\phi_d(t)$ is primarily influenced by pairs of low persistence, whereas when $p$ is large, it is dominated by pairs of high persistence. The function is then vectorized on a uniform grid with resolution $r \times r$. 

    \item Persistence Image: Given a linear transformation $T: \mathbb{R}^2 \to \mathbb{R}^2$, where $T(x, y) = (x, y - x )$. By applying this linear transformation to the points of the persistence diagram we obtain the set:
    
\begin{equation}
T\left(\mathcal{D}_{d}\right)=\left\{ \left(z,w\right)\in\mathbb{R}^{2};\;z=x,\;w=y-x,\;where\;\left(x,y\right)\in\mathcal{D}_{d}\right\} .\label{eq:T-applied-D_d}
\end{equation}

 To convert this new set of points into an image, we use a differentiable probability distribution with a mean vector $\mathbf{u} = (u_x, u_y) \in \mathbb{R}^2$. Specifically, we employ the normalized symmetric Gaussian distribution, $g_u$, which has a mean of $\mathbf{u}$ and a variance of $\sigma^2$, and is defined as follows:

    \begin{equation}
        g_u(x, y) = \frac{1}{2 \pi \alpha^2} e^{-[(x - u_x)^2 + (y-u_y)^2]/2\sigma^2}. \label{eq:gauss-def}
    \end{equation}
    Fix a nonnegative weighting function $f \colon \mathbb{R}^2 \rightarrow \mathbb{R}$ that is zero along the horizontal axis, continuous, and piecewise differentiable. From this, we can obtain a corresponding persistence surface $\sigma_{\mathcal{D}} \colon \mathbb{R}^2 \rightarrow \mathbb{R}$ defined by the following function:
    \begin{equation}
        \sigma_{\mathcal{D}}(z) = \sum_{u \in T(\mathcal{D})}{f(u)g_u(z)}.
        \label{eq:persistence-image3}
    \end{equation}

   Finally, the surface $\sigma_{\mathcal{D}}(z)$ is reduced to a finite-dimensional vector by discretizing a relevant subdomain and integrating $\sigma_{\mathcal{D}}(z)$ over each region in the discretization. To do this, we fix a grid $\mathcal{G}$ in the plane and assign to each pixel $\mathbf{p} \in \mathcal{G}$ the integral of $\sigma_{\mathcal{D}}(z)$ over the corresponding cell in the grid $\mathcal{G}$, given by:
   \begin{equation}
       I(\sigma_{\mathcal{D}})_\mathbf{p} = \int_\mathbf{p} \\ \sigma_{\mathcal{D}} \,dy\,dx
   \end{equation}

    \item Heat Kernel:    Given a persistence diagram $\mathcal{D}_{d} = \{(b_i, d_i)\}_{i=1}^n$, where its points are treated as the locations of Dirac deltas, we can create two distinct vectorizations. One of these vectorizations, known as the symmetry heat vectorization, is formed through the solution of the problem:
    \begin{equation}
        \begin{cases}
        
        \Delta_x u =  \partial_t u,& \text{      in  } \Omega \times \mathbb{R}_{+}^{*},\\
        u = 0, &\text{       on  } \partial \Omega \times \mathbb{R}_{+},\\
        u = \sum_{\mathbf{p} \in \mathcal{D}_{d}} \delta_p,              & \text{        on  } \Omega \times {0},
        \end{cases}
\label{eq:heat-equation-problem}
    \end{equation}
    where $\Omega = \{(x_1, x_2) \in \mathbb{R}^2 ,|, x_1 \geq x_2\}$. We can solve the same equation by first applying a change of coordinates, transforming $(x_1, x_2)$ into $(x_2, x_1)$. Then, we define the image of the set $\mathcal{D}$ in first quadrant $\mathbb{R}_{+}$ as the difference between the solutions obtained using these two coordinate transformations at a specified time $t$. 
    Remembering that the solution of the problem (\ref{eq:heat-equation-problem}), with the initial condition given by a single Dirac delta function supported at a point $\mathbf{p} \in \mathbb{R}^2$, is
    \begin{equation}
    g_p(x) = \frac{1}{4 \pi t}  \exp \left(-\frac{ \|p-x\|^2} {4t} \right). \label{eq:sol-eq-heat-single}
    \end{equation}
        To emphasize the connection with normally distributed random variables, it's customary to employ the change of variable $\sigma = \sqrt{2t}$.  Considering expression (\ref{eq:sol-eq-heat-single}), the solution of problem (\ref{eq:heat-equation-problem}) is:
\begin{equation}
u(x)=\sum_{\mathbf{p}\in\mathcal{D}_{d}}\frac{1}{4\pi t}\exp\left(-\frac{\|p-x\|^{2}}{4t}\right),\label{eq:solution-problem-heat}
\end{equation}

\end{enumerate}

Before extracting the topological features from MNIST images and applying vectorization to the persistence diagrams, they undergo preprocessing, which involves converting grayscale images to binary images and applying three binary image processing functions: heat function, radial function and density function \cite{30}. Remarkably, pixel values vary across different levels, facilitating the capture of information during the filtration process, following the shape of the digits more effectively. The image processing functions will be defined below.

\subsection{Binary Image Processing Functions}
\label{subsec: binary-image}
A $d$-dimensional image $\mathbf{x}\in\mathbb{R}^{n_1 \times n_2 \times \ldots \times n_d}$ can be formalized as a map $\mathcal{I} \colon I \subseteq \mathbb{Z}^d \longrightarrow  \mathbb{R_{+}}$, where each element of $I$ is called a voxel (or pixel when $d = 2$) and $\mathcal{I}(\mathbf{p})$ is the intensity of the image or grayscale value at pixel $\mathbf{p}$. In the same way, a binary image is a map $\mathcal{B} \colon I \subseteq \mathbb{Z}^d \longrightarrow \{0,1\}$, where each pixel $\mathbf{p}$ has intensity write ($\mathcal{I}(\mathbf{p})=1$) or black ($\mathcal{I}(\mathbf{p})=0$).

\subsubsection{Height function}

Let be a binary image $\mathcal{B}\colon I\longrightarrow\{0,1\}$,
where $I\subset\mathbb{Z}^{d}$, and a selected direction $\boldsymbol{v}\in\mathbb{R}^{d}$
with norm $1$. Hence, the height function is defined by $\mathcal{H}_{\boldsymbol{v}}\colon I\times I\longrightarrow\mathbb{R}_{+}$
as: 

\begin{equation}
\mathcal{H}_{\boldsymbol{v}}(\boldsymbol{p}):=\begin{cases}
\langle\boldsymbol{p},\boldsymbol{v}\rangle, & \text{if }\mathcal{B}(\boldsymbol{p})=1,\\
H_{\infty}, & \text{if }\mathcal{B}(\boldsymbol{p})=0,
\end{cases}\label{eq:Height-function}
\end{equation}
where $H_{\infty}$ is the value of the pixel that is the farthest from $v$. 

\subsubsection{Radial function}

Let be a binary image $\mathcal{B}\colon I\longrightarrow\{0,1\}$,
where $I\subset\mathbb{Z}^{d}$, and a selected center $\boldsymbol{c}\in I$.
Then, the radial function is defined by $\mathcal{R}\colon I\longrightarrow\mathbb{R}$,
where $I\subset\mathbb{Z}^{d}$, as follows:

\begin{equation}
\mathcal{R}_{\boldsymbol{c}}(\boldsymbol{p}):=\begin{cases}
\|\boldsymbol{p}-\boldsymbol{c}\|_{2}, & \text{if }\mathcal{B}(\boldsymbol{p})=1,\\
R_{\infty}, & \text{if }\mathcal{B}(\boldsymbol{p})=0,
\end{cases}\label{eq:Radial-function}
\end{equation}
where $R_{\infty}$ is the distance of the pixel that is the farthest
away from the center.

\subsubsection{Density function}

Given $I\subset\mathbb{Z}^{d}$, a binary image $\mathcal{B}\colon I\longrightarrow\{0,1\}$
and a scalar $radious\in\mathbb{R}$, the density function $D_{radious}\colon I\longrightarrow\{0, 1\}$
is defined by:

\begin{equation}
\mathcal{D}_{radious}\left(\boldsymbol{p}\right):=|\{v\in I;\;\mathcal{B}(v)=1\text{ and }\|\boldsymbol{p}-\boldsymbol{v}\|\leq radious\}|,\label{eq:Density-function}
\end{equation}
So $radious$ is a parameter that represents the radius of a ball and $\mathcal{D}_{radious}\left(\boldsymbol{p}\right)$
is the number of neighbors of $\boldsymbol{p}$ with intensity equal
to $1$.

\section{Proposed Method}
\label{sec:proposedmethod}
 
The approach of employing persistence homology with different dimensions and theier persistence diagrams is based on the multivector approach mentioned in \cite{1}, which employs three distinct methods. In the first one,  homology groups with dimensions zero and one are calculated. Then the corresponding persistence diagrams are generated and vetorized separately.\\
\\
The second approach consists of disregarding the dimension of the homology groups. In this case, the persistence diagrams of zero dimension and one dimension are treated as a single diagram, from which a vector representation is calculated. The author of \cite{1} named this approach "fused". Formally, let $\mathcal{D}_{0} = \{(b_i^{0}, d_i^{0})\}_{i=1}^{n_0}$  and $\mathcal{D}_{1} = \{(b_i^{1}, d_i^{1})\}_{i=1}^{n_1}$ be the persistence diagrams then $\mathcal{D}_{fused} = \mathcal{D}_0 \cup \mathcal{D}_1 = \{(b_i^d, d_i^d)_{i=1}^{n}| d = 0, 1\}$.

The third approach is more simplified and involves concatenating the vector representations of the persistence diagrams for each dimension. This approach is referred to as "concat".

For persistence diagrams where the points have a death value of $\infty$, this value is substituted with the highest pixel value in the corresponding scalar fields.

As previously mentioned, the objective of this study is to investigate the use of persistence diagrams in the classification of MNIST images through neural networks. However, before generating the persistence diagrams, we will apply binarization and the functions described in Section \ref{subsec: binary-image}  to each image in the MNIST dataset. 
Through these image processing procedures, the digit pixels are mapped to sublevels, while the image background exhibits higher values. This step simplifies the treatment of the image background in the final stage of the filtration.

The height, radial, and density functions, presented in equations (\ref{eq:Height-function}), (\ref{eq:Radial-function}) and (\ref{eq:Density-function}), respectively, require image binarization. For this purpose, we maintain the threshold of $0.4$, as chosen in \cite{1}, since it preserves the shape of the digits in the images. These functions involve some parameters, such as direction (in the height function), center (in the radial function), and radius (in the density function). We will use eight directions for the height function, which are the vectors $[(0, 1), (0, -1), (1, 0), (-1, 0), (1, 1), (1, -1), (-1, 1), (-1, -1)]$. For the radial function, nine centers will be used given by $[(13, 6), (6, 13), (13, 13), (20, 13), (13, 20), (6, 6), (6, 20), (20, 6), (20, 20)]$, and a radius of $radious = 6$ in the density function. For each image, we will obtain a set of 18 images resulting from the processing using each of the functions (\ref{eq:Height-function})-(\ref{eq:Density-function}) with the just mentioned parameters. Thus, 18 diagrams will be obtained for each dimension for each image.

The vectorization of persistence diagrams is carried out using techniques outlined in Section \ref{sec:mathematicalbackground}. These techniques involve specific parameters that we set as following: 
\begin{itemize}
    \item Betti Curve: The Betti curve necessitates the input parameter resolution, which determines the size of the output vector. In this article, we utilize a resolution of $r = 75$ according to \cite{1}.
    \item Persistence Landscape: The persistence landscape relies on two parameters, resolution and layer numbers. We adopt a resolution of $r = 75$  according to \cite{1} and explore layer numbers within the set $n \in \{1, 3, 5\}$.
    \item Persistence Silhouette: For this technique, only the resolution parameter is required. In our experiments, we set resolution to $r = 75$  according to \cite{1}. We used the parameter power $p = 1$ which is default parameter of the giotto library \cite{20}. 
    \item Persistence Image: In this method, we specify two parameters which are resolution and weight function. We set the resolution to $r = 10$  according to \cite{1}, which defines the grid size for image generation as $r^2$. To compute equation (\ref{eq:persistence-image3}), we employ a constant weight function, $f(x) = 1$ for all $x$, and set the standard deviation to $ \sigma = 0.1$ in expression (\ref{eq:gauss-def}) .
    \item Heat Kernel: In this technique, we use a resolution of $r = 10$ and set the standard deviation to $\sigma = 0.1$ in equation (\ref{eq:heat-equation-problem}).

\end{itemize}

\subsection{MNIST Dataset}
The MNIST database (\textit{Modified National Institute of Standards and Technology database}) consists of images of handwritten digits ranging from 0 to 9, collected from various sources. It is one of the most widely used datasets in the fields of computer vision and machine learning. It plays a pivotal role in image classification algorithms and serves as a valuable resource for demonstrating image processing and deep learning techniques.

Each image in the MNIST dataset has dimensions of $28 \times 28$ pixels and is represented in grayscale. In total, the dataset comprises $70.000$ images, typically employing $60.000$ images for training and $10.000$ images for testing.
In the context of this article, as referenced in \cite{1}, we have chosen to use $5.000$ images for training and $1.250$ images for testing. This choice is aimed at containing the computational costs associated with the process.

\subsection{Proposed Neural Networks}
\label{neural-network}
In the field of neural networks, two of the standout architectures are the Convolutional Neural Networks (CNN) and the Multi-Layer Perceptrons (MLP), due to their versatility and adaptability to a wide range of tasks. In this work, the applied architectures are built based on the MNIST-MLP and MNIST-CNN neural networks, which are traditionally used for image classification in the MNIST dataset, as implemented in the TensorFlow library \cite{21}.

However, in contrast to the conventional inputs used in these TensorFlow architectures, which typically utilize vectorized or non-vectorized MNIST image pixels as input, we also provide the topological characteristics of the images as input. Specifically, through the vectorization of persistence diagrams, it becomes possible to use topological features as input in the MLP architectures while the CNNs still receive only the MNIST images as input.

\begin{figure}
  \centering
  \includegraphics[scale=0.5]{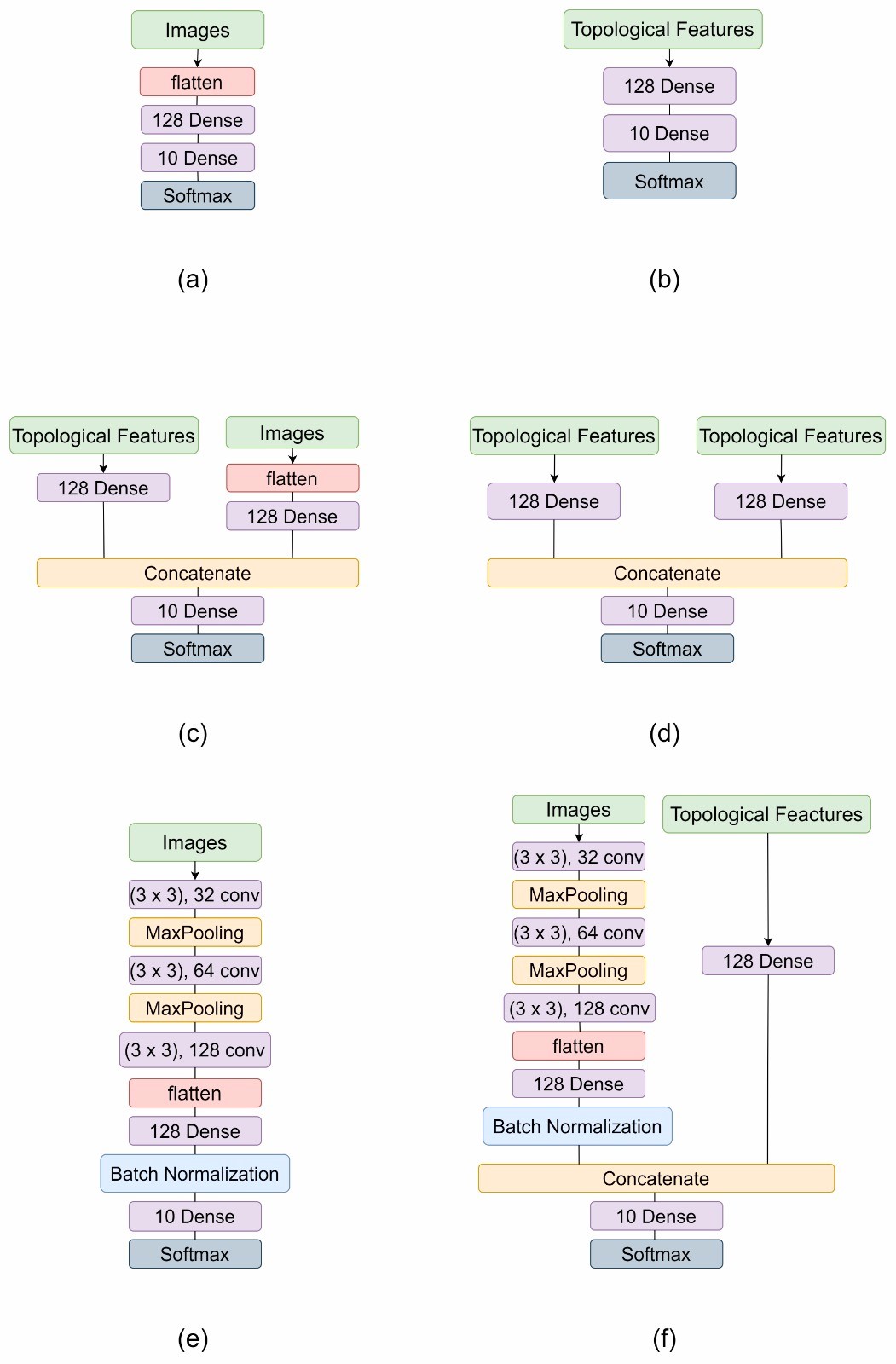}
  \caption{Architectures of neural networks: (a) MNIST-MLP Image; (b) MNIST-MLP Topological; (c) MNIST-MLP Topological +  MNIST-MLP Image; (d) MNIST-MLP Topological + MNIST-MLP Topological; (e)  MNIST-CNN Image; (f) MNIST-CNN Image + MNIST-MLP Topological.}
  \label{fig:fig4}
\end{figure}

    We employ six architectural approaches for image classification, as depicted in Figure \ref{fig:fig4}. These approaches are categorized based on the type of neural network used (MNIST-MLP or MNIST-CNN) and the initials of input they utilize: I (of Image) or T (of Topological features). Since MNIST-CNN receive only images as input, this architecture will be represented only by MNIST-CNN. We hereby introduce and label these architectures as follows:

\begin{enumerate}
    \item MNIST-MLP-I (Figure \ref{fig:fig4}.(a)): MNIST-MLP with original vectorized image as input; 
    \item MNIST-MLP-T (Figure \ref{fig:fig4}.(b)): MNIST-MLP with persistence diagram vectorized as input;

    \item MNIST-MLP-T + MNIST-MLP-I (Figure \ref{fig:fig4}.(c)): two stream architecture that receives persistence diagrams vectorized and original vectorized image.
    \item MNIST-MLP-T + MNIST-MLP-T (Figure \ref{fig:fig4}.(d)): two stream architecture that processes persistence diagrams vectorized.
    \item MNIST-CNN (Figure \ref{fig:fig4}.(e)): MNIST-CNN with original image as input;
    \item MNIST-CNN + MNIST-MLP-T (Figure \ref{fig:fig4}.(f)): two stream architecture that have as input original images and persistence diagram vectorized.

\end{enumerate}

The architecture illustrated in Figure \ref{fig:fig4}.(a) is an MLP (Multi-Layer Perceptron) neural network designed to take a set of images as input during training. This architecture incorporates a flatten layer, responsible for vectorizing each image. Following this, there is a hidden layer with 128 neurons utilizing the Rectified Linear Unit (Relu) activation function. Finally, there is a layer with 10 neurons employing the softmax activation function. This function enforces the neural network's output to represent the probability of data belonging to one of the defined classes.

On the other hand, the architecture in Figure \ref{fig:fig4}.(b) is another MLP neural network, but it takes the topological features of the images as input during training. This architecture also includes a hidden layer with 128 neurons using the Relu activation function and a layer with 10 neurons that utilizes the softmax activation function.

The architecture shown in Figure \ref{fig:fig4}.(c) is a concatenation of the architectures illustrated in Figures \ref{fig:fig4}.(a) and \ref{fig:fig4}.(b). Meanwhile, the architecture in Figure \ref{fig:fig4}.(d) is a concatenation of the architecture in Figure \ref{fig:fig4}.(b) with different topological features as input. Lastly, the architecture depicted in Figure \ref{fig:fig4}.

The architecture depicted in Figure \ref{fig:fig4}.(e) is a convolutional neural network (CNN) designed to process a set of images as input during training. This architecture comprises three Conv2D layers, responsible for conducting convolutions. Each filter (32 in the first convolution layer, 64 in the second, and 128 in the third) transforms the images using a $3 \times 3$ window in all three Conv2D layers. This transformation is applied across the entire image. Additionally, the architecture includes two MaxPooling layers, serving as downsampling filters. These layers reduce a $2 \times 2$ matrix of the image to a single pixel, retaining the maximum value within the $2 \times 2$ matrix. The primary goal of these layers is to preserve the essential features of the image while reducing its dimensions. The architecture also incorporates a Flatten layer, responsible for converting the tensors into a 1D vector. Following that, there is a hidden dense layer with 128 neurons utilizing the Relu activation function. The second-to-last layer in this architecture is Batch Normalization, which normalizes its inputs by applying a transformation that maintains the output mean close to 0 and the output standard deviation close to 1. Lastly, the last layer is a Softmax layer with 10 neurons. Meanwhile, the architecture in Figure \ref{fig:fig4}.(f) is a concatenation of the architectures shown in Figures \ref{fig:fig4}.(b) and \ref{fig:fig4}.(e).

\section{Computational Results}
 The results in this section showcase the performance of the neural network architectures, as depicted in Figure \ref{fig:fig4}, for the classification of MNIST images. Through rigorous 10-fold cross-validation, we calculated both the mean accuracy and standard deviation, as detailed in this section. Notably, the outcomes achieved using the architectures presented in Figure \ref{fig:fig4}.(a) and \ref{fig:fig4}.(c), as summarized in Table \ref{tab:table1}, consistently demonstrate accuracy levels surpassing the 90\% mark. It's worth highlighting that, by incorporating specific topological features, these neural network architectures managed to outperform the traditional MNIST-MLP architecture, which relies on image pixels as input only. These compelling results are further elaborated in Tables \ref{tab:table2}-\ref{tab:table11}, underscoring the potential of leveraging topological features to enhance image classification accuracy. The results that showed a significant increase compared to the MNIST-MLP Image and MNIST-CNN Image architectures are highlighted in orange in the tables.

\label{sec:results}

\begin{table}[h]
 \caption{ Mean accuracy of neural networksMNIST- MNIST MLP and MNIST-MNIST CNN.}
  \centering
  \begin{tabular}{ll}

    \cmidrule(r){2-2}
    & Accuracy   \\
    \midrule
    MNIST-CNN Image  & $0.98768 \pm  0.00536$\\
    MNIST-MLP Image  & $0.94224 \pm 0.01251$    \\
    
    \bottomrule
  \end{tabular}
  \label{tab:table1}
\end{table}

\begin{table}[h]
 \caption{Mean accuracy of MNIST dataset using the PD Betti Curve representation ($r = 75, \sigma= 0.1$).}
  \centering

  \begin{tabular}{lllll}

    \cmidrule(r){2-5}
    &  $H_0$     & $H_1$     & fused     & concat \\
    \midrule

     MNIST-MLP-T & \cellcolor[HTML]{F8A102} $0.94288 \pm 0.01573  $ & $0.48352 \pm 0.02017$   & \cellcolor[HTML]{F8A102}$0.96272 \pm 0.01227$ & \cellcolor[HTML]{F8A102} $0.96448 \pm 0.01338$   \\
    MNIST-MLP-T + MNIST-MLP-I     &\cellcolor[HTML]{F8A102} $0.94464 \pm 0.00944$       & \cellcolor[HTML]{F8A102} $0.94704 \pm 0.00798$ & \cellcolor[HTML]{F8A102}$0.95296 \pm 0.00497$ & \cellcolor[HTML]{F8A102} $0.94720 \pm 0.01054$  \\
    MNIST-CNN + MNIST-MLP-T     & $0.98176 \pm 0.00625$       & $0.97968 \pm 0.00526$ & $0.98496 \pm 0.00373$ & $0.97936 \pm 0.01093$ \\
    \bottomrule
  \end{tabular}
  \label{tab:table2}
\end{table}

\begin{table}[H]
 \caption{Mean accuracy of MNIST dataset using the PD persistence landscape representation ($r = 75, n = 1$)}
  \centering
  \begin{tabular}{lllll}

    \cmidrule(r){2-5}
    &  $H_0$     & $H_1$     & fused     & concat \\
    \midrule
     MNIST-MLP-T & $0.93664 \pm 0.01136$ & $0.67456 \pm 0.01329$   & \cellcolor[HTML]{F8A102}$0.95744 \pm 0.01123$ & \cellcolor[HTML]{F8A102}$0.95408 \pm 0.01258$   \\
    MNIST-MLP-T + MNIST-MLP-I     & \cellcolor[HTML]{F8A102}$0.94336 \pm 0.00737$       & \cellcolor[HTML]{F8A102} $0.949120 \pm 0.00753$ & \cellcolor[HTML]{F8A102} $0.94879 \pm 0.00744$ &\cellcolor[HTML]{F8A102} $0.94784 \pm 0.00709$  \\
    MNIST-CNN + MNIST-MLP-T    & $0.98096 \pm 0.00583$       & $0.98799 \pm 0.00761$ & $0.97904 \pm 0.00552$ & $0.98256 \pm 0.00513$ \\
    \bottomrule
  \end{tabular}
  \label{tab:table3}
\end{table}

\begin{table}[H]
 \caption{Mean accuracy of MNIST dataset using the PD persistence landscape representation ($r = 75, n = 3)$.}
  \centering
  \begin{tabular}{lllll}

    \cmidrule(r){2-5}
    &  $H_0$     & $H_1$     & fused     & concat \\
    \midrule
     MNIST-MLP-T & \cellcolor[HTML]{F8A102} $0.94192 \pm 0.011845 $ & $0.65584 \pm 0.01467$   & \cellcolor[HTML]{F8A102}$0.95216 \pm 0.00892$ & \cellcolor[HTML]{F8A102} $0.94784 \pm 0.01042$   \\
    MNIST-MLP-T + MNIST-MLP-I     & \cellcolor[HTML]{F8A102} $0.94592 \pm 0.009182$       & \cellcolor[HTML]{F8A102} $0.94480 \pm 0.00806$ & \cellcolor[HTML]{F8A102}$0.95088 \pm 0.00611$ & \cellcolor[HTML]{F8A102} $0.94512 \pm 0.00919$  \\
        MNIST-CNN + MNIST-MLP-T     & $0.97568 \pm 0.01151$       & $0.98096 \pm  0.00565$ & $0.98096 \pm 0.00565$ & $0.98112 \pm 0.00796$ \\
    \bottomrule
  \end{tabular}
  \label{tab:table4}
\end{table}

\begin{table}[H]
 \caption{Mean accuracy of MNIST dataset using the PD persistence landscape representation ($r = 75, n = 5)$.}
  \centering
  \begin{tabular}{lllll}
    \cmidrule(r){2-5}
    &  $H_0$     & $H_1$     & fused     & concat \\
    \midrule
     MNIST-MLP-T & $0.92832 \pm 0.01432 $ & $0.65392 \pm 0.01957$   & \cellcolor[HTML]{F8A102}$0.95600 \pm 0.01150$ &\cellcolor[HTML]{F8A102} $0.95632 \pm 0.01015$   \\
    MNIST-MLP-T +  MNIST-MLP-I     & \cellcolor[HTML]{F8A102} $0.94704 \pm 0.01074$       &\cellcolor[HTML]{F8A102} $0.95328 \pm  0.01106$ & \cellcolor[HTML]{F8A102}$0.95434 \pm 0.01083$ & \cellcolor[HTML]{F8A102}$0.95216 \pm 0.01021$  \\
    MNIST-CNN + MNIST-MLP-T     & $0.98352 \pm 0.00469$       & $0.97968 \pm 0.00568$ & $0.98016 \pm 0.00560$ & $0.98416 \pm 0.00466$ \\
    \bottomrule
  \end{tabular}
  \label{tab:table5}
\end{table}

\begin{table}[H]
 \caption{Mean accuracy of MNIST dataset using the PD persistence silhouette representation ($r = 75)$.}
  \centering
  \begin{tabular}{lllll}
    \cmidrule(r){2-5}
    &  $H_0$     & $H_1$     & fused     & concat \\
    \midrule
     MNIST-MLP-T & $0.67536 \pm 0.00801$ & $0.45104 \pm 0.01980$   & $0.82752 \pm 0.01571$ & $0.78192 \pm  0.013596$   \\
    MNIST-MLP-T + MNIST-MLP-I     & $0.93872 \pm 0.00882$       & \cellcolor[HTML]{F8A102} $0.94720 \pm 0.00936$ & \cellcolor[HTML]{F8A102} $0.94240 \pm 0.01009 $  & \cellcolor[HTML]{F8A102} $0.94448 \pm  0.00632$  \\
    MNIST-CNN + MNIST-MLP-T     & $0.98288 \pm 0.00757$       & $0.98496 \pm 0.00481$ & $0.98592 \pm 0.00334$ & $0.98400 \pm 0.00516$ \\
    \bottomrule
  \end{tabular}
  \label{tab:table6}
\end{table}

\begin{table}[H]
 \caption{Mean accuracy of MNIST dataset using the PD persistence image representation ($r = 10)$.}
  \centering
  \begin{tabular}{lllll}
    \cmidrule(r){2-5}
    &  $H_0$     & $H_1$     & fused     & concat \\
    \midrule
     MNIST-MLP-T & $0.91264 \pm 0.01812 $ & $0.28784 \pm 0.02180$   & \cellcolor[HTML]{F8A102} $0.95568 \pm 0.01386$ & $0.89712 \pm 0.01129$   \\
    MNIST-MLP-T + MNIST-MLP-I     & \cellcolor[HTML]{F8A102}$0.94496 \pm 0.00748$       & $0.93840 \pm 0.00758$ & \cellcolor[HTML]{F8A102} $0.95040 \pm 0.01101$ & \cellcolor[HTML]{F8A102} $0.94576 \pm 0.00982$  \\
    MNIST-CNN + MNIST-MLP-T    & $0.98064 \pm 0.00823$       & $0.98384 \pm 0.00513$ & $0.97248 \pm 0.01245$ & $0.98016 \pm 0.00424$ \\
    \bottomrule
  \end{tabular}
  \label{tab:table7}
\end{table}

\begin{table}[H]
 \caption{Mean accuracy of MNIST dataset using the PD heat kernel representation ($r = 10)$.}
  \centering
  \begin{tabular}{lllll}
    \cmidrule(r){2-5}
    &  $H_0$     & $H_1$     & fused     & concat \\
    \midrule
     MNIST-MLP-T &\cellcolor[HTML]{F8A102} $0.96512 \pm 0.03998$ & $0.66240 \pm 0.02602$   & \cellcolor[HTML]{F8A102}$0.98416 \pm 0.02421$ & \cellcolor[HTML]{F8A102}$0.97984 \pm 0.02922$   \\
    MNIST-MLP-T + MNIST-MLP-I     & $0.93648 \pm 0.00586$       &\cellcolor[HTML]{F8A102} $0.95136 \pm 0.01118$ & $0.93824 \pm 0.01012$ & $0.94048 \pm 0.01063$  \\
    MNIST-CNN + MNIST-MLP-T     & $0.95376 \pm 0.01069$       & $0.98159 \pm 0.00481$ & $0.96272 \pm 0.00719$ & $0.95936 \pm 0.00791$ \\
    \bottomrule
  \end{tabular}
  \label{tab:table8}
\end{table}

\begin{table}[H]
 \caption{Mean accuracy of MNIST dataset using the concatente of Betti curve $(r = 75)$ and persistence landscape $(r = 75, n = 3)$.}
  \centering
  \begin{tabular}{lllll}
    \cmidrule(r){2-5}
    &  $H_0$     & $H_1$     & fused     & concat \\
    \midrule
     MNIST-MLP-T + MNIST-MLP-T &\cellcolor[HTML]{F8A102}  $0.94400 \pm 0.01208$ & $0.66304 \pm 0.02755$   &\cellcolor[HTML]{F8A102} $0.9568 \pm 0.01302$ & \cellcolor[HTML]{F8A102}$0.95936 \pm 0.01167$   \\

    \bottomrule
  \end{tabular}
  \label{tab:table9}
\end{table}

\begin{table}[H]
 \caption{Mean accuracy of MNIST dataset using the concatenate of Betti curve $(r = 75)$ and persistence silhouette $(r = 75)$.}
  \centering
  \begin{tabular}{lllll}
    \cmidrule(r){2-5}
    &  $H_0$     & $H_1$     & fused     & concat \\
    \midrule
     MNIST-MLP-T + MNIST-MLP-T & $0.94208 \pm 0.00736$ & $ 0.66128 \pm 0.01823$  &  \cellcolor[HTML]{F8A102} $0.95760 \pm 0.00573$ & \cellcolor[HTML]{F8A102} $0.95408 \pm 0.01052$   \\

    \bottomrule
  \end{tabular}
  \label{tab:table10}
\end{table}

\begin{table}[H]
 \caption{Mean accuracy of MNIST dataset using the concatenate of persistence landscape $(r = 75, n = 3)$ and persistence silhouette $(r = 75)$.}
  \centering
  \begin{tabular}{lllll}
    \cmidrule(r){2-5}
    &  $H_0$     & $H_1$     & fused     & concat \\
    \midrule
     MNIST-MLP-T + MNIST-MLP-T & $0.93952 \pm 0.01082$ & $0.65952 \pm 0.02729$  & \cellcolor[HTML]{F8A102} $0.94799 \pm 0.01174$ & \cellcolor[HTML]{F8A102} $0.95312 \pm 0.01104$   \\

    \bottomrule
  \end{tabular}
  \label{tab:table11}
\end{table}

Note that heat kernel vectorization in Table \ref{tab:table8} showed higher accuracies surpassed the MLP Image architecture. However, persistence silhouette vectorization in Table \ref{tab:table6} did not yield a significant increase in accuracy for both the MNIST-CNN and MNIST-MLP-I architectures. The other vectorization showed accuracies increase for the MLP Image architecture, but do not surpass the heat kernel in the MLP Topological and MNIST-MLP-T + MNIST-MLP-I, as reported in Tables \ref{tab:table2} - \ref{tab:table4}, \ref{tab:table7}.

The tables from \ref{tab:table9} to \ref{tab:table11} that utilize a combination of persistence diagram vectorization methods have shown notable improvements in accuracy compared to the MNIST-MLP-I architecture. However, it's worth noting that this increase in performance is not significantly large compared to architectures that use only one of the vectorization methods.

Table \ref{tab:table12} presents the best results from architectures that used topological features as input, while Table \ref{tab:table13} displays the accuracies in 10-fold cross-validation of results of the Table \ref{tab:table12}.

\begin{table}[H]
 \caption{Better results of MNIST dataset.}
  \centering
  \begin{tabular}{llll}
    \cmidrule(r){1-4}
      Homology     & Accuracy     & Vectorization &    Architecture \\
    \midrule
     $H_0$ & $0.96512$ & Heat Kernel & MNIST-MLP-T    \\
    $H_1$ & $0.95136$ & Heat Kernel  & MNIST-MLP-T + MNIST-MLP-I    \\
    fused & $0.98416$ & Heat Kernel  & MNIST-MLP-T  \\
    concat     & $0.97984$ & Heat Kernel  & MNIST-MLP-T   \\
    \bottomrule
  \end{tabular}
  \label{tab:table12}
\end{table}

\begin{table}[H]
 \caption{
Accuracy of the K-fold cross-validation with the best results.}
  \centering
  \begin{tabular}{lllll}
    \cmidrule(r){1-5}
   Accuracy &  $H_0$     & $H_1$     & fused     & concat \\
    \midrule
     Run 1     & $0.97920$ & $0.95680$ & $0.99520$ & $0.99040$   \\
     Run 2     & $0.99360$ & $0.95360$ & $0.99680$ & $0.99520$  \\
     Run 3     & $0.97280$ & $0.95680$ & $0.99680$ & $0.99520$ \\
     Run 4     & $0.99040$ & $0.94560$ & $0.99680$ & $0.99520$ \\
     Run 5     & $0.98240$ & $0.93440$ & $0.99999$ & $0.99520$ \\
     Run 6     & $0.99200$ & $0.93440$ & $0.99680$ & $0.99040$ \\
     Run 7     & $0.97920$ & $0.95040$ & $0.99360$ & $0.99520$ \\
     Run 8     & $0.97440$ & $0.94240$ & $0.99840$ & $0.99360$ \\
     Run 9     & $0.87680$ & $0.94400$ & $0.92320$ & $0.91200$ \\
     Run 10    & $0.88640$ & $0.94560$ & $0.93440$ & $0.90720$ \\
             
    \bottomrule

  \end{tabular}
  \label{tab:table13}
\end{table}

\section{Conclusions and Future Works}
\label{sec:conclusions}
Our approach encompassed both MNIST-CNN and MNIST-MLP architectures, exploring both topological features and image pixels as inputs. The obtained results highlight the potential of the proposed neural network architectures in the context of MNIST image classification. 

It is noteworthy that incorporating topological features as inputs in certain CNN architectures did not yield significant improvements in accuracy compared to those using image pixels as inputs. However, in contrast to the CNN architecture, discernible improvements were observed when the MLP used topological features as inputs.

It is important to note that the use of topological features in neural networks does not prove advantageous in terms of computational cost but rather in terms of result accuracy. Despite an increase in computational processing, topological features demonstrated the potential to enhance accuracy in multi-class image classification tasks.

Those conclusions suggest the necessity for critical analysis of results, including those related to dimension one that did not yield favorable results, and encourage further studies involving changes in the neural network architectures, such as exploring the number of layers.
\\
\\
\\
\\
\\
\\
\\
\\
\\
\\
\\
\\
\bibliographystyle{unsrt}  

\end{document}